%% file: main.tex
\documentclass{article}\usepackage{beamerarticle}
\usepackage[numbers]{natbib}

\input{commands.tex}
\mode<presentation> {

\usetheme{Frankfurt}

\usecolortheme{orchid}

}

\usepackage{graphicx} 
\usepackage{booktabs} 
\usepackage{amsmath} 
\usepackage{amsfonts}
\usepackage{amssymb}
 \newcommand{\VE}{\ensuremath{\mathcal{V}}}
\newcommand{\Vf}{\ensuremath{\mathcal{V}^f_z}}
\newcommand{\Vz}{\ensuremath{\mathcal{V}_z}}

\newcommand{\CE}{\ensuremath{\mathcal{C}}}
\newcommand{\DE}{\ensuremath{\mathcal{D}}}
\usepackage{authblk}
\usepackage{fancyhdr}
\title[Learning and VE]{One Formalization of  Virtue Ethics via Learning}

\author{Naveen Sundar Govindarajulu$^*$}
\author{Selmer Bringjsord}
\author{Rikhiya Ghosh}
\affil{Rensselaer AI \& Reasoning Lab, \\Rensselaer Polytechnic
  Institue (RPI),
Troy, New York 12180, USA\\
$^*$E-mail: {govinn2}@rpi.com\\
www.rpi.edu
}
 
\date{\today} 

\begin{document}
\maketitle

\begin{abstract}
\noindent
Given that there exist many different formal and precise treatments of
deontological and consequentialist ethics, we turn to virtue ethics
and consider what could be a formalization of virtue ethics that makes
it amenable to automation.  We present an embroyonic formalization in
a cognitive calculus (which subsumes a quantified first-order logic)
that has been previously used to model robust ethical principles, in
both the deontological and consequentialist traditions.
\end{abstract}
\begin{footnotesize}
\tableofcontents
\end{footnotesize}
\newpage
\section{Introduction}

Separate from the two main camps in ethics, \textbf{deontological
  ethics} ($\DE$) and \textbf{consequentialism} (\CE), there is
\textbf{virtue ethics} (\VE).  While there has been extensive formal,
computational, and mathematical work done on deontological ethics and
consequentialism, there has been very little or almost no work done in
formalizing and making rigorous virtue ethics. Proponents of \VE\
might claim that it is not feasible to do so given \VE 's emphasis on
character and traits, rather than individual actions or consequencens.
From the perspective of machine and robot ethics, this is not
satisfactory. If \VE\ is to be considered to be on equal footing with
\DE\ and \CE\ for the purpose of building morally competent machines,
we need to start with formalizing parts of virtue ethics.  (After all,
machines don't yet understand that which is informal; witness e.g.\
SIRI.)  We present one such formalization based on learning and using
one version of virtue ethics presented by Zagzebski in
\cite{zagzebski2010exemplarist}. The goal in this paper is to present
a simple formalization of a virtue ethics theory in a formal calculus
that has been used to model deontological and consequentialist
principles \cite{nsg_sb_dde_2017,dde_self_sacrifice_2017}.


The plan for the paper is as follows.  First, we present a very quick
overview of virtue ethics.  Then we cover related work that can be
considered as formalizations of virtue ethics.  We then present one
version of virtue ethics, $\VE_z$, that we seek to formalize fully.
Then our calculus and the formalization itself (\Vf) are presented.
We conclude by discussing future work and challenges.

 \section{An Overview of Virtue Ethics}  
 In simple forms of \CE, actions are evaluated based on their total
 utlity to everyone involved.  The best action is the action that has
 the highest total utility. In \DE, the emphasis is on inviolable
 principles, and reasoning from those principles to whether actions
 are obligatory, permissible, neutral, etc.  In contrast to \DE\ and
 \CE, some forms of virtue ethics can be summed up by saying the best
 action in a situation is the action that a virtuous person would do.
 A virtuous person is defined as a person that has learnt and
 internalized a diverse set of virtuous habits or traits.  For a
 virtuous person, virtuous acts become second-nature, and hence are
 performed in many different situations.  Note that unlike \DE\ and
 \CE, it is not entirely straightforward how one could translate these
 notions into a form that is precise enough to be realized in
 machines.


\section{Related Prior Work}
\begin{figure}[h!]
 \centering
 {
  \includegraphics[width=0.45\linewidth]{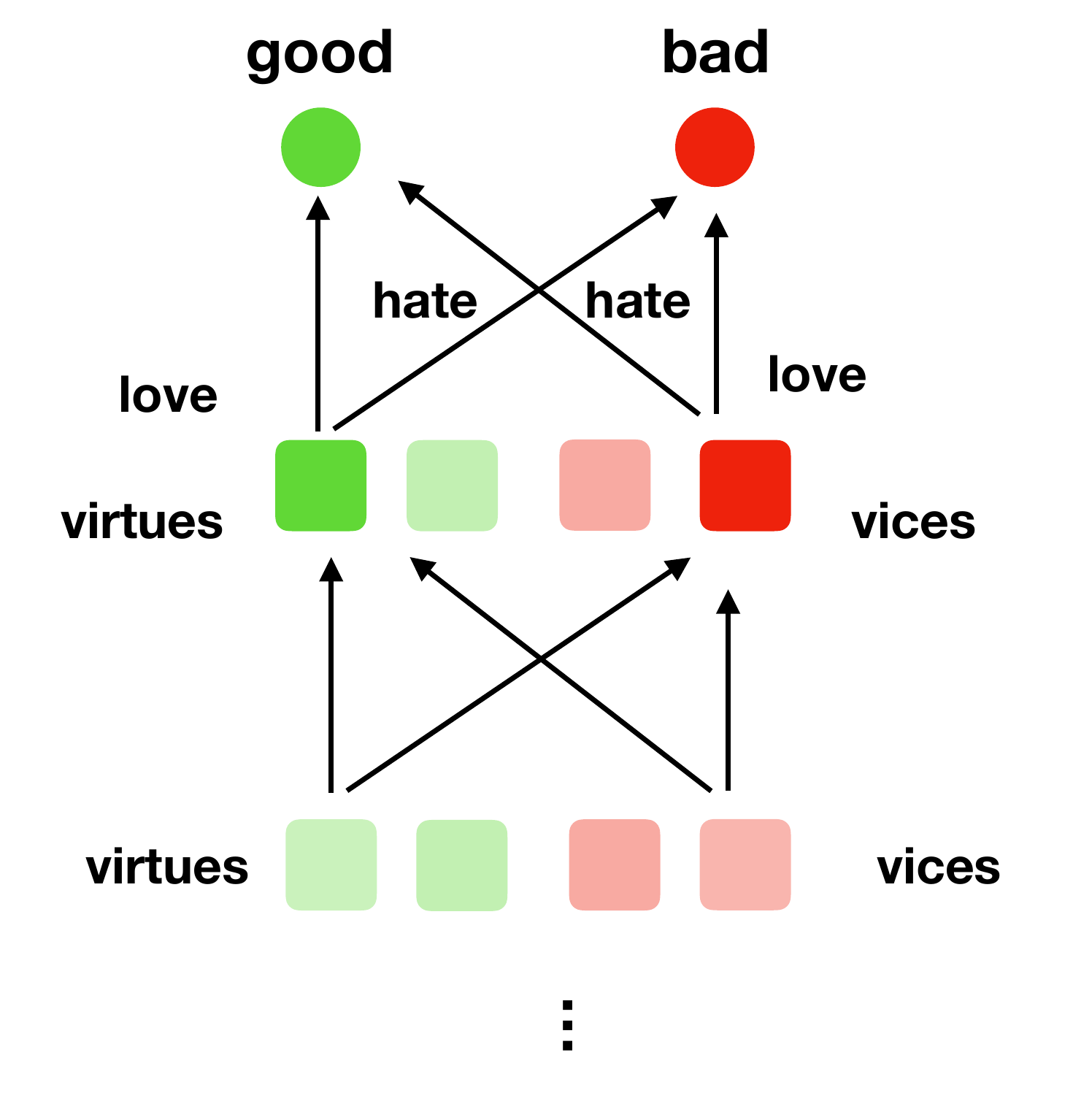}}
 \caption{Hurka's account}
 \label{fig:ccOverview}
\end{figure}

Hurka \cite{hurka2000virtue} presents an ingenious formal account
involving a recursive notion of goodness and badness. Hurka starts
with a given set of primitive good and bad states of affairs. Virtues
are then defined as love of good states of affairs or hatred of bad
states of affairs. Vice is defined as love of bad states of affairs or
hatred of good states of affairs. Virtues and vices are then
themselves taken to be good and bad states of affairs, resulting in a
recursive definition (See Figure~\ref{fig:ccOverview}). While there are rectifiable issues
\cite{hiller2011unusual} with Hurka's recursive definition, we feel
that Hurka's definition might not capture central aspects of virtue
\cite{miles2013against}.  We feel that it still has to be shown that
this account is different from rigorous and formal accounts of
\CE. Moreover, it is not clear how this account can be exploited for
automation (Note: This is not Hurka' goal).

\section{Exemplarist Virtue Theory}

\input{evt.tex}

\section{The Goal}

\input{goal.tex}

\section{Building the Formal Machinery}

 \label{sect:calc}
\input{calculus.tex}

\subsection{Formalizing Emotions}
To formalize emotions, we build upon the \textbf{OCC model}.
\label{sect:occ}
\input{occ.tex}

\label{sect:inf}

\input{emotions.tex}

\subsection{Learning Method} 
Note that when we look at humans learning virtues by observing others
or by reading from texts or other sources, it is not entirely clear
how models of learning that have been successful in perception and
language processing (e.g. the recent successes of deep
learning/differentiable learning/statistical learning) can be be
applied. Learning in these situations is from one or few instances or
in some cases through instruction and such learning may not be readily
amenable to models of learning which require a large number of
examples.

The abstract learning method that we will use is
\textbf{generalization}. If we have a set of set of formulae
$\{\Gamma_1, \ldots, \Gamma_n\}$, the generalization of
$\{\Gamma_1, \ldots, \Gamma_n\}$, denoted by $g\big(\{\Gamma_1, \ldots, \Gamma_n\}\big)$ is a $\Gamma$ such that
$\Gamma\vdash \wedge \Gamma_i$. See one simple example below:

\begin{mdframed}[linecolor=white, frametitle=Example 1,
  frametitlebackgroundcolor=gray!10, backgroundcolor=gray!05,
  roundcorner=8pt]
  \begin{equation*}
 \begin{aligned}
&\Gamma_1  = \{ \mathit{talkingWith} (\mathit{jack}) \rightarrow \mathit{Honesty} \}\\
&\Gamma_2  = \{ \mathit{talkingWith} (\mathit{jill}) \rightarrow \mathit{Honesty} \}\\
\midrule
\mathbf{generalization } \  &\Gamma  = \{ \forall x. \mathit{talkingWith}(x) \rightarrow  \mathit{Honesty} \}
\end{aligned}
\end{equation*}
\end{mdframed}

One particularly efficient and well-studied mechanism to realise
generalization is \textbf{anti-unification}. Anti-unification that has
been applied successfully in learning programs from few
examples.\footnote{This discipline known as inductive programming seeks
  to build precise computer programs from examples \cite{nienhuys1997foundations}. } In anti-unification, we are given a set of
expressions $\{f_1, \ldots, f_n\}$ and we need to compute an
expression $g$ that when substituted with an appropriate term
$\theta_i$ gives us $f_i$. E.g. if we are given
$\mathit{hungry}(\mathit{jack})$ and $\mathit{hungry}(\mathit{jill})$,
the anti-unification of those terms would be $\mathit{hungry}(\mathit{x})$.

\begin{mdframed}[linecolor=white, frametitle=Example 2,
  frametitlebackgroundcolor=gray!10, backgroundcolor=gray!05,
  roundcorner=8pt]
  \begin{equation*}
 \begin{aligned}
&\mathit{likes}(\mathit{jill, jack})\\
&\mathit{likes}(\mathit{jill, jim})\\
\midrule
\mbox{\textbf{anti-unification }} \ & \mathit{likes}(\mathit{jill, x})
\end{aligned}
\end{equation*}
\end{mdframed}

In higher-order anti-unification, we can substitute function symbols
and predicate symbols. Here $P$ is a higher-order variable.

\begin{mdframed}[linecolor=white, frametitle=Example 3,
  frametitlebackgroundcolor=gray!10, backgroundcolor=gray!05,
  roundcorner=8pt]
  \begin{equation*}
 \begin{aligned}
&\mathit{likes}(\mathit{jill, jack})\\
&\mathit{loves}(\mathit{jill, jim})\\
\midrule
\mbox{\textbf{anti-unification }} \ & \mathit{P}(\mathit{jill, x})
\end{aligned}
\end{equation*}
\end{mdframed}
 
\subsection{Defining Traits }
We need agents to learn traits and not just single actions. We define
below what it means for an agent to have a trait. First,  a situation
$\sigma(t)$ is simply a collection of formulae that describes what
fluents hold at a time $t$ along with other event calculus constraints
and descriptions. An action type $\alpha$ is said to consistent in a
situation $\sigma(t)$ for an agent $a$ if:

$$\sigma(t)+
\happens\big(\action(\alpha, a), t\big) \not\vdash \bot$$

\begin{mdframed}[linecolor=white, frametitle=Trait,
  frametitlebackgroundcolor=gray!25, backgroundcolor=gray!10,
  roundcorner=8pt]
An agent
$a$ is said to have an action type $\alpha$ as a trait if there are at
least $m$
situations $\{ \sigma_1 , \sigma_2 , \ldots, \sigma_n \}$ in which
there are unique alternatives $\{\alpha_1, \ldots, \alpha_m\}$ available but
\emph{instantiations} of $\alpha$ is performed in a large fraction $\gamma\gg 1 $of these situations. 

\end{mdframed}

\subsection{Learning from Exemplars and Not Just From Examples}
We start with a learning agent $l$. An agent $e$ is identified as an
exemplar by $l$ \emph{iff} the corresponding emotion of admiration is
triggered $n$ times or more. A learnt trait is defined below:

\begin{mdframed}[linecolor=white, frametitle=Learnt Trait,
  frametitlebackgroundcolor=gray!25, backgroundcolor=gray!10,
  roundcorner=8pt]
  A learnt trait is simply a situation $\sigma(t)$ and an action type
  $\alpha$: $\langle \sigma(t), \alpha \rangle$
\end{mdframed}
Once $e$ is identified, the learner then identifies one or more traits
of $e$ by observing $e$ over an extended period of time. Let
$\{ \sigma_1, \sigma_2, \ldots, \sigma_n\}$ be the set of situations
in which instantiations $\{ \alpha_1, \alpha_2, \ldots, \alpha_n\}$ of
a particular trait $\alpha$ are triggered. The learner then simply
associates the action type $\alpha$ with the generalization of the situations
$g(\{ \sigma_1 , \sigma_2, \ldots, \sigma_n\})$. That
is the agent has incorporated this learnt trait:

$$\Big\langle g\big(\{ \sigma_1, \sigma_2, \ldots, \sigma_n\}\big),
\alpha \Big\rangle$$
For instance, if the trait is \emph{``being truthful''} and is triggered in
situations: \emph{``talking with alice,''}, \emph{``talking with
  bob''}, \emph{``talking with charlie''}; then the association learnt
is that \emph{``talking with an agent''} should trigger the
\emph{``being truthful''} action type.

\section{Example} 

We present a simple example. Assume that we have a market place where
things that are broken or unbroken can be bought and sold. A seller
can either honestly state the condition of the item
$\{\mathit{broken}, \mathit{unbroken}\}$ or not correctly report the
state of the item. For an honest seller, we have the following two
situations that can be observed:
 
\begin{mdframed}[linecolor=white, frametitle=Situation 1,
  frametitlebackgroundcolor=gray!05, backgroundcolor=gray!05,
  roundcorner=8pt]
    \begin{equation*}
 \begin{aligned}
\sigma_1 &\equiv \holds(broken,t) \\
\alpha & \equiv \happens(utter(broken), t)\\
  \end{aligned}
\end{equation*}
\end{mdframed}

\begin{mdframed}[linecolor=white, frametitle=Situation 2,
  frametitlebackgroundcolor=gray!05, backgroundcolor=gray!05,
  roundcorner=8pt]
    \begin{equation*}
 \begin{aligned}
 \sigma_2 &\equiv \holds(unbroken,t) \\
\alpha & \equiv \happens(utter(unbroken), t)
 \end{aligned}
\end{equation*}
\end{mdframed}
The learnt trait is then given below. The trait says that one should
always correctly utter the state of the item.

\begin{mdframed}[linecolor=white,  frametitlebackgroundcolor=gray!05, backgroundcolor=gray!05,
  roundcorner=8pt]
    \begin{equation*}
 \begin{aligned}
 \Big\langle \holds(x,t),  \happens(utter(x), t) \Big\rangle
 \end{aligned}
\end{equation*}
\end{mdframed}

\section{Conclusion}
We have presented an initial formalization of a virtue ethics theory
in a calculus that has been used in automating other ethical
principles in deontological and consequentialist ethics. Many
important questions have to be addressed in future research. Among
them, are questions about the nature and source of the utility
functions that are used in the definitions of emotions. We also need
to apply this model to realistic examples and case studies. The lack
of such formal examples and case studies is a bottleneck here.

\bibliographystyle{plain}
\bibliography{main72,naveen}

\end{document}

%% file: commands.tex
\usepackage{times}
 \usepackage{soul}
\usepackage{paralist}

 \usepackage[small]{caption}
\usepackage{amsfonts}
\usepackage{amsmath}
\usepackage{mathtools}    
\usepackage{relsize}
\usepackage{hyperref}
\usepackage{proof}  
\usepackage{bussproofs}

\usepackage{pifont}
\usepackage{cleveref}

\newcommand{\IGNORE}[1]{}

\newcommand{\lsort}[1]{%
  \ensuremath{\mbox{\textsf{#1}}}}
\newcommand{\defsort}[2]{%
  \newcommand{#1}{\lsort{#2}}}
\defsort{\Action}{Action}
\defsort{\Time}{Time}
\defsort{\Self}{Self}

\defsort{\Agent}{Agent}
\defsort{\Entrant}{Entrant}
\defsort{\ActionType}{ActionType}
\defsort{\Moment}{Moment}
\defsort{\Boolean}{Formula}
\defsort{\PayOut}{PayOut}
\defsort{\Fluent}{Fluent}
\defsort{\Event}{Event}
\defsort{\Object}{Object}
\defsort{\RealTerm}{RealTerm}

\defsort{\Numeric}{Numeric}
\defsort{\Number}{Number}
\defsort{\Trolley}{Trolley}
\defsort{\Track}{Track}
\defsort{\Moveable}{Moveable}

\newcommand{\lsymbol}[1]{%
  \ensuremath{\mathit{#1}}}
\newcommand{\defsymbol}[2]{%
  \newcommand{#1}{\lsymbol{#2}}}
\defsymbol{\actionType}{action}
\defsymbol{\initially}{initially}
\defsymbol{\holds}{holds}
\defsymbol{\happens}{happens}
\defsymbol{\clipped}{clipped}
\defsymbol{\initiates}{initiates}
\defsymbol{\terminates}{terminates}
\defsymbol{\prior}{prior}
\defsymbol{\interval}{interval}

\defsymbol{\does}{does}
\defsymbol{\plans}{plans}
\defsymbol{\act}{act}
\defsymbol{\react}{react}
\defsymbol{\payTot}{pay_{tot}}
\defsymbol{\fight}{fight}
\defsymbol{\coop}{coop}
\defsymbol{\enter}{enter}
\defsymbol{\stayout}{stayout}
\defsymbol{\learns}{learns}
\defsymbol{\payoff}{payoff}
\defsymbol{\position}{position}
\defsymbol{\dead}{dead}
\defsymbol{\damaged}{damaged}

\defsymbol{\onrails}{onrails}
\defsymbol{\switch}{switch}
\defsymbol{\drop}{drop}

\newcommand{\lconstant}[1]{%
  \ensuremath{\mbox{\textsf{#1}}}}
\newcommand{\defconstant}[2]{%
  \newcommand{#1}{\lconstant{#2}}}
\defconstant{\Enter}{Enter}
\defconstant{\StayOut}{StayOut}
\defconstant{\Fight}{Fight}
\defconstant{\Acquiesce}{Acquiesce}
\defconstant{\cs}{cs }

\newcommand{\lmodality}[1]{%
  \ensuremath{\mathbf{#1}}}
\newcommand{\defmodality}[2]{%
  \newcommand{#1}{\lmodality{#2}}}
\defmodality{\knows}{K}
\defmodality{\believes}{B}
\defmodality{\perceives}{P}
\defmodality{\mental}{M}

\defmodality{\desires}{D}
\defmodality{\intends}{I}
\defmodality{\says}{S}
\defmodality{\ought}{O}

\newcommand{\sep}{\ \lvert \ }

\usepackage{booktabs}
\usepackage{mdframed}
\usepackage{paralist}

\usepackage{color}

\newcommand{\DDE}{\ensuremath{\mathcal{{DDE}}}}
\newcommand{\DCEC}{\ensuremath{{\mathcal{{DCEC}}}}}

\newcommand{\type}[1]{\textsf{#1}}

\providecommand{\shortcite}[1]{\cite{#1}}

\newcommand{\Distress}{\ensuremath{\mathsf{Distress}}}
\newcommand{\HappyFor}{\ensuremath{\mathsf{HappyFor}}}
\newcommand{\PityFor}{\ensuremath{\mathsf{PityFor}}}
\newcommand{\Gloating}{\ensuremath{\mathsf{Gloating}}}
\newcommand{\Resentment}{\ensuremath{\mathsf{Resentment}}}

\newcommand{\joy}{\ensuremath{\mathit{joy}}}
\newcommand{\distress}{\ensuremath{\mathit{distress}}}
\newcommand{\happyFor}{\ensuremath{\mathit{happyFor}}}

\newcommand{\admires}{\ensuremath{\mathit{admires}}}

%% file: evt.tex
\textbf{Exemplarist virtue theory} (\Vz) builds on the \textbf{direct
  reference theory} (DRT) of semantics and has the emotion of
\textbf{admiration} as a foundational object. In DRT, the meaning of a
word is constructed by what the word points out. For example, to
understand the meaning of \emph{``water''}, a person need not
understand and possess all knowledge about water. The person simply
needs to understand that ``water'' points to something which is
similar to \emph{that} (with \emph{that} pointing to water).

In \Vz, moral terms are assumed to be understood similarly. Moral
attributes are defined by direct reference when instantiated in
exemplars (saints, sages, heroes) that one identifies through
admiration.  The emotions of admiration and contempt play a
foundational role in this theory. Zagzebski posits a process very
similar to scientific or empirical investigation, Exemplars are first
identified and their traits are studied. Exemplars are then
continously further studied to better understand their traits,
qualities, etc. The status of an individual as an exemplar can change
over time. Below is an informal version that we seek to formalize:

\begin{small}

\begin{mdframed}[frametitle={ Informal Version \Vz},linewidth=0,backgroundcolor=gray!7,frametitlebackgroundcolor=gray!15]

\begin{enumerate}
\item[$\mathbf{I}_1$] Agent or person $a$ perceives a person $b$
  perform an action $\alpha$. This observation causes the emotion of
  admiration in $a$

\item[$\mathbf{I}_2$]   $a$ then studies $b$ and seeks to
  learn what traits (habits/dispositions) $b$ has.
\end{enumerate}

\end{mdframed}
\end{small}


%% file: goal.tex
From the above presentation of \Vz, we can glean the following
distilled requirements that should be present in any formalization.

\begin{small}

\begin{mdframed}[frametitle={Requirements},linewidth=0,backgroundcolor=gray!7,frametitlebackgroundcolor=gray!15]

\begin{enumerate}
\item[$\mathbf{R}_1$] A formalization of emotions, particularly admiration.

\item[$\mathbf{R}_2$] A notion of learning traits (and not just simple individual actions).
\end{enumerate}

\end{mdframed}
\end{small}


%% file: calculus.tex
The computational logic we use is the \textbf{deontic cognitive event
  calculus} (\DCEC).  This logic was used previously in
\cite{nsg_sb_dde_2017,dde_self_sacrifice_2017} to automate versions of
the doctrine of double effect \DDE, an ethical principle with
deontological and consequentialist components.  While describing the
calculus is beyond the scope of this paper, we give a quick overview
of the system.  Dialects of \DCEC\ have also been used to formalize
and automate highly intensional reasoning processes, such as the
false-belief task \cite{ArkoudasAndBringsjord2008Pricai} and
\textit{akrasia} (succumbing to temptation to violate moral
principles) \cite{akratic_robots_ieee_n}. {Arkoudas and Bringsjord
  \shortcite{ArkoudasAndBringsjord2008Pricai} introduced the general
  family of \textbf{cognitive event calculi} to which \DCEC\ belongs,
  by way of their formalization of the false-belief task.} \DCEC\ is a
sorted (i.e.\ typed) quantified modal logic (also known as sorted
first-order modal logic) that includes the event calculus, a
first-order calculus used for commonsense reasoning.  The calculus has
a well-defined syntax and proof calculus; see Appendix A of
\cite{nsg_sb_dde_2017}. The proof calculus is based on natural
deduction \cite{gentzen_investigations_into_logical_deduction}, and
includes all the introduction and elimination rules for first-order
logic, as well as inference schemata for the modal operators and
related structures.

\subsection{Syntax}
\label{subsect:syntax}

As metioned above, \DCEC\ is a sorted calculus.  A sorted system can
be regarded analogous to a typed single-inheritance programming
language.  We show below some of the important sorts used in \DCEC.\\
 
\begin{scriptsize}
\begin{tabular}{lp{5.8cm}}  
\toprule
Sort    & Description \\
\midrule
\type{Agent} & Human and non-human actors.  \\

\type{Time} &  The \type{Time} type stands for
time in the domain.  E.g.\ simple, such as $t_i$, or complex, such as
$birthday(son(jack))$. \\

 \type{Event} & Used for events in the domain. \\
 \type{ActionType} & Action types are abstract actions.  They are
  instantiated at particular times by actors.  Example: eating.\\
 \type{Action} & A subtype of \type{Event} for events that occur
  as actions by agents. \\
 \type{Fluent} & Used for representing states of the world in the
  event calculus. \\
\bottomrule
\end{tabular}
\end{scriptsize} \\

The syntax has two components: a first-order
core and a modal system that builds upon this first-order core.  The
figures below show the syntax and inference schemata of \DCEC.    The first-order core of \DCEC\ is
the \emph{event calculus} \cite{mueller_commonsense_reasoning}.
Commonly used function and relation symbols of the event calculus are
included.  Fluents, event and times are the three major sorts of the event
calculus. Fluents represent states of the world as first-order
terms. Events are things that happen in the world at specific instants
of time. Actions are events that are carried out by an agent. For any
action type $\alpha$ and agent $a$, the event corresponding to $a$
carrying out $\alpha$ is given by $action(a, \alpha)$. For instance
if $\alpha$ is \textit{``running''} and $a$ is \textit{``Jack'' },
$action(a, \alpha)$ denotes \textit{``Jack is running''}.
Other calculi (e.g.\ the \emph{situation calculus}) for
modeling commonsense and physical reasoning can be easily switched out
in-place of the event calculus.

 \begin{scriptsize}
\begin{mdframed}[linecolor=white, frametitle=Syntax,
  frametitlebackgroundcolor=gray!25, backgroundcolor=gray!10,
  roundcorner=8pt]
 \begin{equation*}
 \begin{aligned}
    \mathit{S} &::= 
    \begin{aligned}
      & \Agent \sep \ActionType \sep \Action \sqsubseteq
      \Event \sep \Moment  \sep \Fluent \\
    \end{aligned} 
    \\ 
    \mathit{f} &::= \left\{
    \begin{aligned}
      & action: \Agent \times \ActionType \rightarrow \Action \\
      &  \initially: \Fluent \rightarrow \Boolean\\
      &  \holds: \Fluent \times \Moment \rightarrow \Boolean \\
      & \happens: \Event \times \Moment \rightarrow \Boolean \\
      & \clipped: \Moment \times \Fluent \times \Moment \rightarrow \Boolean \\
      & \initiates: \Event \times \Fluent \times \Moment \rightarrow \Boolean\\
      & \terminates: \Event \times \Fluent \times \Moment \rightarrow \Boolean \\
      & \prior: \Moment \times \Moment \rightarrow \Boolean\\
    \end{aligned}\right.\\
        \mathit{t} &::=
    \begin{aligned}
      \mathit{x : S} \sep \mathit{c : S} \sep f(t_1,\ldots,t_n)
    \end{aligned}
    \\ 
    \mathit{\phi}&::= \left\{ 
    \begin{aligned}
     & q:\Boolean \sep  \neg \phi \sep \phi \land \psi \sep \phi \lor
     \psi \sep \forall x: \phi(x) \sep \\\
 &\perceives (a,t,\phi)  \sep \knows(a,t,\phi) \sep     \\ 
& \common(t,\phi) \sep
 \says(a,b,t,\phi) 
     \sep \says(a,t,\phi) \sep  \believes(a,t,\phi) \\
& \desires(a,t,\phi)  \sep \intends(a,t,\phi) \\ & \ought(a,t,\phi,(\lnot)\happens(action(a^\ast,\alpha),t'))
      \end{aligned}\right.
  \end{aligned}
\end{equation*}
\end{mdframed}
\end{scriptsize}
The modal operators present in the calculus include the standard
operators for knowledge $\knows$, belief $\believes$, desire
$\desires$, intention $\intends$, etc.  The general format of an
intensional operator is $\knows\left(a, t, \phi\right)$, which says
that agent $a$ knows at time $t$ the proposition $\phi$.  Here $\phi$
can in turn be any arbitrary formula. Also,
note the following modal operators: $\mathbf{P}$ for perceiving a
state, 
$\mathbf{C}$ for common knowledge, $\mathbf{S}$ for agent-to-agent
communication and public announcements, $\mathbf{B}$ for belief,
$\mathbf{D}$ for desire, $\mathbf{I}$ for intention, and finally and
crucially, a dyadic deontic operator $\mathbf{O}$ that states when an
action is obligatory or forbidden for agents. It should be noted that
\DCEC\ is one specimen in a \emph{family} of easily extensible
cognitive calculi.
 
The calculus also includes a dyadic (arity = 2) deontic operator
$\ought$. It is well known that the unary ought in standard deontic
logic lead to contradictions.  Our dyadic version of the operator
blocks the standard list of such contradictions, and
beyond.\footnote{A overview of this list is given lucidly in
  \cite{sep_deontic_logic}.}

\subsection{Inference Schemata}

The figure below shows a fragment of the inference schemata for \DCEC.
First-order natural deduction introduction and elimination rules are
not shown. Inference schemata $R_\mathbf{K}$ and $R_\mathbf{B}$ let us
model idealized systems that have their knowledge and beliefs closed
under the \DCEC\ proof theory.  While humans are not dedcutively
closed, these two rules lets us model more closely how more deliberate
agents such as organizations, nations and more strategic actors
reason. (Some dialects of cognitive calculi restrict the number of
iterations on intensional
operators.) 
$R_4$ states that knowledge of a proposition implies that the
proposition holds $R_{13}$ ties intentions directly to perceptions
(This model does not take into account agents that could fail to carry
out their intentions).  $R_{14}$ dictates how obligations get
translated into known intentions.

\begin{scriptsize}

\begin{mdframed}[linecolor=white, frametitle=Inference Schemata
  (Fragment), nobreak=true, frametitlebackgroundcolor=gray!25, backgroundcolor=gray!10, roundcorner=8pt]
\begin{equation*}
\begin{aligned}
  &\hspace{40pt} \infer[{[R_{\knows}]}]{\knows(a,t_2,\phi)}{\knows(a,t_1,\Gamma), \ 
    \ \Gamma\vdash\phi, \ \ t_1 \leq t_2}  \\ 
& \hspace{40pt} \infer[{[R_{\believes}]}]{\believes(a,t_2,\phi)}{\believes(a,t_1,\Gamma), \ 
    \ \Gamma\vdash\phi, \ \ t_1 \leq t_2} \\
& \hspace{20pt} \infer[{[R_4]}]{\phi}{\knows(a,t,\phi)}
\hspace{18pt}\infer[{[R_{13}]}]{\perceives(a,t', \psi)}{t<t', \ \ \intends(a,t,\psi)}\\
&\infer[{[R_{14}]}]{\knows(a,t,\intends(a,t,\chi))}{\begin{aligned}\ \ \ \ \believes(a,t,\phi)
 & \ \ \
 \believes(a,t,\ought(a,t,\phi, \chi)) \ \ \ \ought(a,t,\phi,
 \chi)\end{aligned}}
\end{aligned}
\end{equation*}
\end{mdframed}
\end{scriptsize}

\subsection{Semantics}

The semantics for the first-order fragment is the standard first-order
semantics. The truth-functional connectives
$\land, \lor, \rightarrow, \lnot$ and quantifiers $\forall, \exists$
for pure first-order formulae all have the standard first-order
semantics. The semantics of the modal operators differs from what is
available in the so-called Belief-Desire-Intention (BDI) logics
{\cite{bdi_krr_1999}} in many important ways.  For example, \DCEC\
explicitly rejects possible-worlds semantics and model-based
reasoning, instead opting for a \textit{proof-theoretic} semantics and
the associated type of reasoning commonly referred to as
\textit{natural deduction}
\cite{gentzen_investigations_into_logical_deduction,proof-theoretic_semantics_for_nat_lang}.
Briefly, in this approach, meanings of modal operators are defined via
arbitrary computations over proofs, as we will see for the
counterfactual conditional below.

 
%
%

%
%

%% file: occ.tex
There are many models of emotion from psychology and cognitive
science. Among these, the OCC model \cite{occ_main} has found wide
adoption among computer scientists. Note that the model presented by
\cite{occ_main} is informal in nature and one formalization of
the model has been presented in \cite{adam2009logical}. The
formalization by \cite{adam2009logical} is based on
propositional modal logic, and while comprehensive and elaborate, is
not expressive enough for our modelling, which requires at the least
a quantified modal logic.

In OCC, emotions are short-lived entities that arise in response to
\emph{events}. Different emotions arise based on whether the
\emph{consequences} to events are positive (desirable) or negative
(undesirable), whether the event has occured, whether the event has
consequences for the agent or for another agent. OCC assumes an
undefined primitive notion of an agent being \emph{pleased} or
\emph{displeased} in response to an event. We represent this notion by
$\Theta$ defined later. Though OCC has twenty two emotion types, we
consider only the following handful of emotions shown below. An agent
can be pleased or displeased in response to an event's consequences
that hold either for the agent or for another agent.  The following
table summarizes the OCC definitions for six emotion types that we
deem hold immediately for us here.

\begin{center}
\begin{footnotesize}
\begin{tabular}{llcl}  
\toprule
\textbf{Emotion Type} & \textbf{Response} &  \textbf{Agent}  & \textbf{Consequences}  \\
\midrule
\ Joy & Pleased & Self & Desirable \\
 \Distress & Displeased & Self & Undesirable \\
\midrule
 \HappyFor & Pleased & Other & Desirable \\
\Gloating  & Pleased & Other & Undesirable \\
 \PityFor & Displeased & Other & Undesirable \\
 \Resentment & Displeased & Other & Desirable \\
\bottomrule
\end{tabular}
\end{footnotesize}
\end{center}




%% file: emotions.tex
Central to the formalization of below is a utility function $\mu$
that maps fluents and time points to utility values.

\begin{footnotesize}
\begin{equation*}
\begin{aligned}
\mu: \Fluent \times \Time \rightarrow \mathbb{R}
\end{aligned}
\end{equation*}
\end{footnotesize} 

The above agent-neutral function suffices for classical \DDE\ but is
not enough for our purpose.  We assume that there is a another
function $\nu$ (either acquired or given to us) that gives us
agent-specific utilities.

\begin{footnotesize}
\begin{equation*}
\begin{aligned}
\nu:  \Agent \times \Fluent  \times \Time \rightarrow \mathbb{R}
\end{aligned}
\end{equation*}
\end{footnotesize} We can then build the agent-neutral function $\mu$ from the
agent-specific function $\nu$ as shown below:

\begin{footnotesize}
\begin{equation*}
\begin{aligned}
\mu(f,t) = \sum_{a}\nu (a, f, t)
\end{aligned}
\end{equation*}
\end{footnotesize}

 \noindent For an event $e$ that happens at time $t$, let
$e_I^{t}$ be the set of fluents initiated by the event, and let
$e_T^{t}$ be the set of fluents terminated by the event.  If
we are looking up till horizon some $H$, then $\bar{\mu}(e, t)$,
the total utility of event $e$ happening at time $t$,
is then:

\begin{footnotesize}
\begin{equation*}
\begin{aligned}
\bar{\mu}(e, t)  = \mathlarger{\mathlarger{‎‎\sum}}_{y=t+1}^{H‎}\Bigg(\sum_{f\in{e_{I}^{t}}} \mu(f, y) - \sum_{f\in{e_{T}^{t}}} \mu(f, y)\Bigg)
\end{aligned}
\end{equation*}
\end{footnotesize}

Similarly, we have $\bar{\nu}(a, e, t)$, the total utility for
agent $a$ of event $e$ that happens at time $t$:

\begin{footnotesize}
\begin{equation*}
\begin{aligned}
\bar{\nu}(a, e, t)  =
\mathlarger{\mathlarger{‎‎\sum}}_{y=t+1}^{H‎}\Bigg(\sum_{f\in{e_{I}^{t}}}
{\nu} (a, f, y) - \sum_{f\in{e_{T}^{t}}} {\nu} (a, f, y)\Bigg)
\end{aligned}
\end{equation*}
\end{footnotesize}

Emotions are fluents comprised of \begin{inparaenum}[(i.)] \item one
  or more agents; \item an event; and \item the time at which event
  took place. \end{inparaenum} 

We now define emotion fluents in terms of the machinery we have
defined above. The general template for an emotion is that an emotion
is equivalent to conditions stated in the OCC formalization and an
addition $\Theta$ condition that is specific to the agent and
time. Not all agents respond in the same way emotionally to the same
conditions. $\Theta$ takes care of this variation. If $\Theta$ always
holds, then we have an agent that is easily swayed by events. If
$\Theta$ never holds, we have an agent that is never emotional. (Free
variables in the following definitions are considered to be
universally quantified.) To set the stage for defining admiration
below, we present the following straightforward definitions for the
four simplest emotions in OCC theory.

\begin{description}
\item[Joy] The agent believes that the total utility of the event for
  itself is positive and that there no negative consequences.

\begin{footnotesize}
\begin{equation*}
\begin{aligned}
  &\ \  \ \ \ \  \ \  \ \  \ \ \ \  \ \  \ \ \ \  \  \ \ \ \holds(\joy(a, e, t), t')\\
 & \ \  \ \ \ \  \ \  \ \  \ \ \ \  \ \  \ \ \ \  \  \ \ \  \ \ \ \ \
 \ \ \ \ \  \ \  \leftrightarrow\\
 & \left[
   \begin{aligned} &  \ \ \  \ \  \ \  \ \ \ \ \ \  \ \  \ \  \ \Theta(a,t') \land \\ 
     \ &\believes\left(a, t', 
       \left[\begin{aligned} &\bar{\nu}(a, e,  t) > 0 \ \  \land \\ &\lnot \exists f,t'. 
           \left(\begin{aligned}
               & initiates(e, f, t) \land \\ &\nu (a, f, t') <
        0 \end{aligned}\right)
     \end{aligned}\right]\right) 
    \end{aligned}\right]
\end{aligned}
\end{equation*}
\end{footnotesize}

\item[Distress] The agent believes that the total utility of the event for
  itself is negative and that there no positive consequences.

\begin{footnotesize}
\begin{equation*}
\begin{aligned}
  &\ \  \ \ \ \  \ \  \ \  \ \ \ \  \ \  \ \ \ \  \  \ \ \ \holds(\distress(a, e, t), t')\\
 & \ \  \ \ \ \  \ \  \ \  \ \ \ \  \ \  \ \ \ \  \  \ \ \  \ \ \ \ \
 \ \ \ \ \  \ \  \leftrightarrow\\
 & \left[
   \begin{aligned} &  \ \ \  \ \  \ \  \ \ \ \ \ \  \ \  \ \  \ \Theta(a,t') \land \\ 
     \ &\believes\left(a, t', 
       \left[\begin{aligned} &\bar{\nu}(a, e,  t) < 0 \ \  \land \\ &\lnot \exists f,t'. 
           \left(\begin{aligned}
               & initiates(e, f, t) \land \\ &\nu (a, f, t') >
        0 \end{aligned}\right)
     \end{aligned}\right]\right) 
    \end{aligned}\right]
\end{aligned}
\end{equation*}
\end{footnotesize}

\item[Happy For] The agent $a$ believes that the total utility of the event for
  agent $b$ is positive and that there no negative consequences. The
  agent also believes that it is different from $b$

\begin{footnotesize}
\begin{equation*}
\begin{aligned}
  &\ \  \ \ \ \  \ \  \ \  \ \ \ \  \ \  \ \ \ \  \  \ \ \ \holds(\happyFor(a,b, e, t), t')\\
 & \ \  \ \ \ \  \ \  \ \  \ \ \ \  \ \  \ \ \ \  \  \ \ \  \ \ \ \ \
 \ \ \ \ \  \ \  \leftrightarrow\\
 & \left[\begin{aligned} &  \ \ \  \ \  \ \  \ \ \ \ \ \  \ \  \ \  \
     \Theta(a,t') \land \\ \ &\believes\left(a, t',
       \left[\begin{aligned} & (a \not= b) \land \bar{\nu}(b, e, t) > 0 \ \  \land \\ &\lnot \exists f,t'. \left(\begin{aligned}
       & initiates(e, f, t) \land \\ &\nu (b, f, t') <
        0 \end{aligned}\right)\end{aligned}\right]\right) \end{aligned}\right]
\end{aligned}
\end{equation*}
\end{footnotesize}

\item[Admiration For]In standard OCC, an agent $a$ is said to admire another agent $b$'s
action $\alpha$, if agent $a$ believes the action is a good action.

\begin{mdframed}[linecolor=white, backgroundcolor=green!05,
  roundcorner=8pt]
\begin{footnotesize}
\begin{equation*}
\begin{aligned}
  &\ \  \ \ \ \  \ \  \ \  \ \ \ \  \ \  \ \ \ \  \  \ \ \ \holds(\admires(a,b, \alpha, t), t')\\
 & \ \  \ \ \ \  \ \  \ \  \ \ \ \  \ \  \ \ \ \  \  \ \ \  \ \ \ \ \
 \ \ \ \ \  \ \  \leftrightarrow\\
 & \left[\begin{aligned} &  \ \ \  \ \  \ \  \ \ \ \ \ \  \ \  \ \  \
     \Theta(a,t') \land \\ \ &\believes\left(a, t',
       \left[\begin{aligned} & (a \not= b) \land \bar{\mu}(
           \actionType(\alpha, b), t) > 0 \ \  \land \\ &\lnot \exists f,t'. \left(\begin{aligned}
       & initiates(\actionType(\alpha, b), f, t) \land \\ &\mu ( f, t') <
        0 \end{aligned}\right)\end{aligned}\right]\right) \end{aligned}\right]
\end{aligned}
\end{equation*}
\end{footnotesize}

\end{mdframed}
\end{description}

